\documentclass{article}
\usepackage{ijcai26}

\usepackage{times}
\usepackage{soul}
\usepackage{url}
\usepackage[hidelinks]{hyperref}
\usepackage[utf8]{inputenc}
\usepackage[small]{caption}
\usepackage{graphicx}
\usepackage{amsmath}
\usepackage{amsthm}
\usepackage{booktabs}
\usepackage{algorithm}
\usepackage{algorithmic}
\usepackage[switch]{lineno}

\title{HighFM: Towards a Foundation Model for Learning Representations from High-Frequency Earth Observation Data}

\author{
Stella Girtsou$^{1,2*}$
\and
Konstantinos Alexis$^{3,4*}$
\and
Giorgos Giannopoulos$^{1}$
\And
Charalambos Kontoes$^{1}$\\
\affiliations
$^1$National Observatory of Athens\\
$^2$National Technical University of Athens\\
$^3$National and Kapodistrian University of Athens\\
$^4$Athena Research Center\\
$^*$Equal contribution\\
\emails
\{sgirtsou@noa.gr, kogalexis@athenarc.gr,
giannopoulos@noa.gr,
kontoes@noa.gr\}
}

\begin{document}

\maketitle

\begin{abstract}
    The increasing frequency and severity of climate-related disasters have intensified the need for real-time monitoring, early warning, and informed decision-making. Earth Observation (EO), powered by satellite data and Machine Learning (ML), offers powerful tools to meet these challenges. Foundation Models (FMs) have revolutionized EO ML by enabling general-purpose pretraining on large-scale remote sensing datasets. However most existing models rely on high-resolution satellite imagery with low revisit rates—limiting their suitability for fast-evolving phenomena and time-critical emergency response. In this work, we present HighFM, a first cut approach towards a FM for high-temporal-resolution, multispectral EO data. Leveraging over 2 TB of SEVIRI imagery from the Meteosat Second Generation (MSG) platform, we adapt the SatMAE masked autoencoding framework to learn robust spatiotemporal representations. To support real-time monitoring, we enhance the original architecture with fine-grained temporal encodings to capture short-term variability. The pretrained models are then fine-tuned on cloud masking and active fire detection tasks. We benchmark our SEVIRI-pretrained Vision Transformers against traditional baselines and recent geospatial FMs, demonstrating consistent gains across both balanced accuracy and IoU metrics. Our results highlight the potential of temporally dense geostationary data for real-time EO, offering a scalable path toward foundation models for disaster detection and tracking.
\end{abstract}

\section{Introduction}

Climate-driven disasters such as wildfires, floods, and extreme storms are accelerating the demand for intelligent, near-real-time Earth Observation (EO) systems capable of supporting rapid environmental decision-making. Monitoring such dynamic environmental phenomena requires EO solutions that go beyond traditional static analyses and provide both high temporal resolution and robust, generalizable models.

Foundation Models (FM) are steadily becoming the state of the art ground model for performing Machine Learning (ML) tasks in various domains including  Natural Language Processing (NLP), Computer Vision (CV), Earth Observation (EO) and also multimodal learning. These models are trained on large-scale, unlabeled datasets using self-supervised strategies, enabling them to learn general purpose representations that can be fine-tuned for a wide range of downstream tasks with minimal supervision. In fact, this self-supervised paradigm has been shown to outperform classical supervised methods in various domains. Notable examples include GPT-4~\cite{openai2024gpt4technicalreport} for language modeling, MAE~\cite{mae2021} for vision and CLIP~\cite{radford2021learningtransferablevisualmodels} for text–image understanding.

The massive volumes of free high resolution satellite images available during the last decades, such as Sentinel mission, has led to the development of a large number of FMs for EO data, as cataloged in recent surveys \cite{awais2023foundationalmodelsdefiningnew,a10916803a,huo2025remotesensing}. Yet, a common limitation remains: nearly all existing EO FMs rely on high-resolution, low-revisit satellite imagery—such as Sentinel-2 and Landsat and are typically evaluated on static or slowly evolving tasks, including land cover classification or phenology estimation. These models, while powerful in detail-rich contexts, are inherently unsuited for rapidly evolving scenarios, such as wildfires, storm development, or dynamic cloud systems.


Geostationary satellite platforms, like Meteosat Second Generation (MSG), Meteosat Third Generation (MTG), GOES and Himawari provide continuous coverage of the Earth’s disk, with revisit times as short as 5 to 15 minutes and spatial resolutions ranging from 500 meters to 3 kilometers depending on the sensor. These platforms are already widely used in operational settings. For example, the FireHUB system~\cite{kontoes2016firehub} employs images from Spinning Enhanced Visible and Infrared Imager (SEVIRI) for real-time wildfire monitoring in Greece; and the Copernicus Atmosphere Monitoring Service (CAMS)~\cite{cams2021forecasts} incorporates aerosol forecasts and cloud products derived from SEVIRI to support solar radiation and air quality modeling. The launch of the new MTG further enhances these capabilities with improved spatial, temporal and spectral resolution in relation to MSG, enabling finer detection of rapidly developing and potentially hazardous weather phenomena.

To bridge the gap between temporally sparse FMs and the demands of real-time EO monitoring, we investigate the development of a FM explicitly designed for high-frequency, geostationary EO data streams. In particular, we focus on the SEVIRI sensor onboard the MSG platform, which provides consistent 15-minute observations across Europe, Africa, and the Middle East. While geostationary sensors provide limited spatial detail, their high revisit rate makes them suitable for tracking fast-evolving phenomena.
To this end, we leverage a self-supervised masked autoencoding framework and pretrain on more than 2 TB of multi-band SEVIRI imagery spanning several years across the Mediterranean. Our approach builds upon the SatMAE architecture but introduces key adaptations to handle the specific characteristics of geostationary data—most notably, the retention of fine-grained temporal encodings typically discarded in slower-evolving EO contexts. We pretrain two variants, using either a single timestep or a three-timestep input and then fine-tune them on fast-evolving downstream tasks; cloud segmentation and pixel-level active fires detection. Across multiple years of evaluation data, the resulting models show strong and consistent performance.
While our focus in this paper is on these two tasks, the underlying model architecture and training methodology are broadly applicable.  The proposed approach, HighFM, establishes a scalable framework for other near-real-time EO applications, including cloud tracking, storm evolution, and solar energy forecasting.

Our key contributions include:

\begin{itemize}
    \item An adapted SatMAE architecture with enhanced temporal encoding for dynamic environmental modeling. To the best of our knowledge, this is the first work introduces the need for FMs specifically tailored for high-frequency EO data.
    \item A curated, large-scale SEVIRI dataset for self-supervised pretraining and two smaller datasets of cloud and fire masks for fine-tuning.
    \item First cut benchmarks on cloud and fire detection across multiple years, showing state-of-the-art performance against strong baselines in both recall- and precision-optimized training objectives.
\end{itemize}

This study is developed in collaboration with operational stakeholders, including the fire brigade, national civil protection service, the national meteorological institute, and scientists working on solar-energy forecasting. This co-design process informs task selection, acceptable error trade-offs (recall versus spatial precision) and evaluation protocols. 

\section{Related work}
Satellites in orbit generate vast volumes of EO data on a daily basis. Due to the scale and complexity of these datasets, comprehensive manual labeling is infeasible, making supervised learning approaches difficult to scale. FMs, which are designed to learn from large amounts of unlabeled data through self-supervised training, are therefore a natural fit for Remote Sensing. This has led to a surge in FM development within the EO domain. Most of these models have been applied and evaluated on core tasks such as land cover classification, object detection, and semantic segmentation.

Self-supervised learning (SSL) has become increasingly popular in Earth Observation (EO). SSL approaches adapted for EO data can be categorized into: (a) Masked Image Modeling (MIM), which reconstructs masked regions to learn spatial–spectral context (e.g., \cite{mae2021,cong2022satmae,bountos2025fomo}); (b) Similarity-based pretraining, which pulls positive (often spatiotemporally related) pairs together and pushes negatives apart; (e.g., \cite{tian2024swimdiff,wang2024softcontrastive,diao2025ringmo}); and (c) Generative modeling, such as Denoising Diffusion Probabilistic Models (DDPMs), used for EO tasks such as cloud removal, missing-data imputation, and super-resolution. (e.g. \cite{khanna2024diffusionsat,tang2024crsdiffcontrollableremotesensing}.

The vast majority of EO FMs have been based on masking models. SatMAE~\cite{cong2022satmae} is one of the first frameworks based on masked autoencoders (MAE)~\cite{mae2021} to tackle EO particularities, including multi-spectral and spatiotemporal location embeddings. It has been pretrained on NASA’s Harmonized Landsat-Sentinel-2 (HLS) dataset, which consists of multi-spectral and temporal satellite imagery and is tested on Land Cover Classification, multi-label classification and building segmentation. Many works have extended original SatMAE; Scale-MAE~\cite{2023scalemae} encodes the resolution of the input image to learn the reconstruction of images at lower/higher scales. FG-MAE~\cite{wang2023featureguidedmaskedautoencoder} extends the standard MAE framework by using remote sensing image features as the reconstruction target, training the model to recover high-level representations rather than raw pixel values.

Considerable progress has also been made in the area of multi-modality, where models are designed to ingest and process heterogeneous datasets from different sensor types (e.g., optical, SAR, physically-based models), as well as varying spatial and temporal resolutions.

OFA-Net~\cite{xiong2024allunifiedfoundationmodels} introduces a unified foundation model using a shared Vision Transformer backbone to pretrain on EO data from diverse modalities and spatial resolutions, including Sentinel-1/-2, Gaofen, NAIP, and EnMAP~\cite{chabrillat2024enmap}. Modality-specific patch embedding layers handle differences in input channels (e.g., 2 bands for Sentinel-1 SAR, 224 for EnMAP hyperspectral). The shared Transformer processes embedded patches across modalities, learning a generalized, robust representation. Training relies on masked image modeling with modality-specific decoders, allowing self-supervised learning without requiring spatial alignment between modalities.

Prithvi~\cite{prithvi} extends MAE to multispectral, multi-temporal EO by using 3D sine–cosine positional encodings (spatial + temporal) and 3D convolutions over spatiotemporal cubes, with a temporal tubelet size of 1 to match low EO revisit rates. It is pretrained on HLS and evaluated on tasks including multi-temporal cloud gap imputation, flood mapping, wildfire scar mapping, and crop segmentation. In contrast, DeCUR~\cite{wang2024decour} targets multimodal self-supervision by decoupling shared from modality-specific representations, improving transfer across heterogeneous sensors (e.g., SAR and optical) when pretrained on datasets such as BigEarthNet~\cite{Sumbul_2019} and SEN12MS~\cite{schmitt2019sen12mscurateddataset}.

While the field is rapidly advancing toward more generalized, multimodal, and spatiotemporal foundation models (FMs), the vast majority of existing FMs are trained on high-resolution satellite imagery with long revisit times which limits their utility for real-time monitoring tasks. SatVision-TOA~\cite{spradlin2024satvisiontoageospatialfoundationmodel} is an early step for higher-frequency EO foundation models, pretrained on MODIS images and evaluated on 3D cloud retrieval; we extend this direction by targeting near-real-time, temporally dense geostationary data for rapid  monitoring.

\section{Methodology}

Our goal is to lay the foundations for the development of a FM capable of real-time or near-real-time environmental monitoring. 
Geostationary platforms (such as MSG, MTG and GOES) are particularly well-suited for such tasks due to their high temporal sampling frequency over large areas. Although their spatial resolution is relatively low compared to polar-orbiting satellites, the frequent revisit times enable effective tracking of diurnal cycles and fast-changing environmental conditions. To assess the quality of the produced models, we focus on two rapidly evolving pixel-wise segmentation tasks: active fire detection and cloud segmentation.

Building on these advantages of geostationary observations, we leverage the high temporal frequency and large volume of available SEVIRI data to construct a large pretraining dataset. Using a self-supervised learning approach with a masked autoencoding strategy, our model learns general spatiotemporal patterns from multi-channel satellite inputs without labeled data. This masking strategy is also naturally aligned with Earth Observation, where cloud cover, smoke, or sensor noise can result in missing or occluded pixels. We then fine-tune the pretrained models for active fire detection and cloud segmentation using curated datasets of SEVIRI image patches paired with corresponding fire and cloud masks.

\begin{figure}[h!]
    \centering
    \includegraphics[width=0.6\columnwidth]{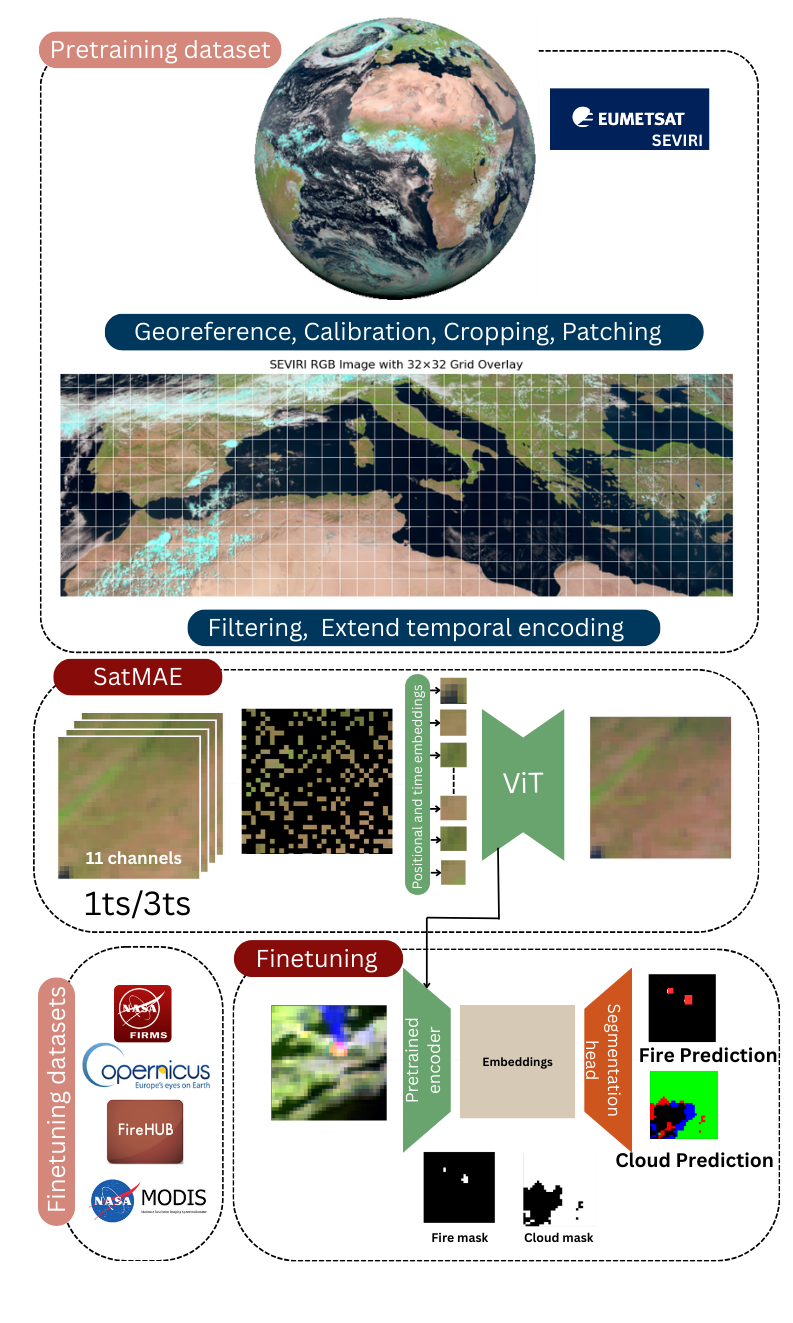}
    \caption{High level flowchart of our methodology}
    \label{fig:fig1}
\end{figure}

\subsection{Datasets}

\noindent \textbf{Pretraining dataset.} The pretraining dataset was built from MSG/SEVIRI radiance imagery, which provides multispectral observations at 15-minute cadence covering a field-of-view of approximately ±80. We use 11 spectral bands (excluding HRV) over the Mediterranean for the period 2014–2019. SEVIRI cloud mask products are used only for data curation and quality control (not as supervision). To align with operational wildfire monitoring, we restrict the archive to May–September, corresponding to the peak Mediterranean fire season.

We implemented a preprocessing pipeline to harmonize the radiances and masks, subset scenes to the Mediterranean region, and generate the associated timestamps required for temporal encoding. Each scene is then split into non-overlapping 32×32 patches, discarding patches that contain only ocean or are fully cloud-covered. The final cleaned pretraining dataset totals approximately 2.23 TB.
To ensure robust model evaluation and avoid temporal data leakage, we adopt a strict time-based split: 2014–2018 are used for pretraining, while 2019 is reserved for evaluation and partitioned across validation and test sets. The validation split is used to monitor pretraining and select hyperparameters, and the test split remains fully held out for final reporting.

\begin{figure}[h!]
    \centering
    \includegraphics[width=0.6\columnwidth]{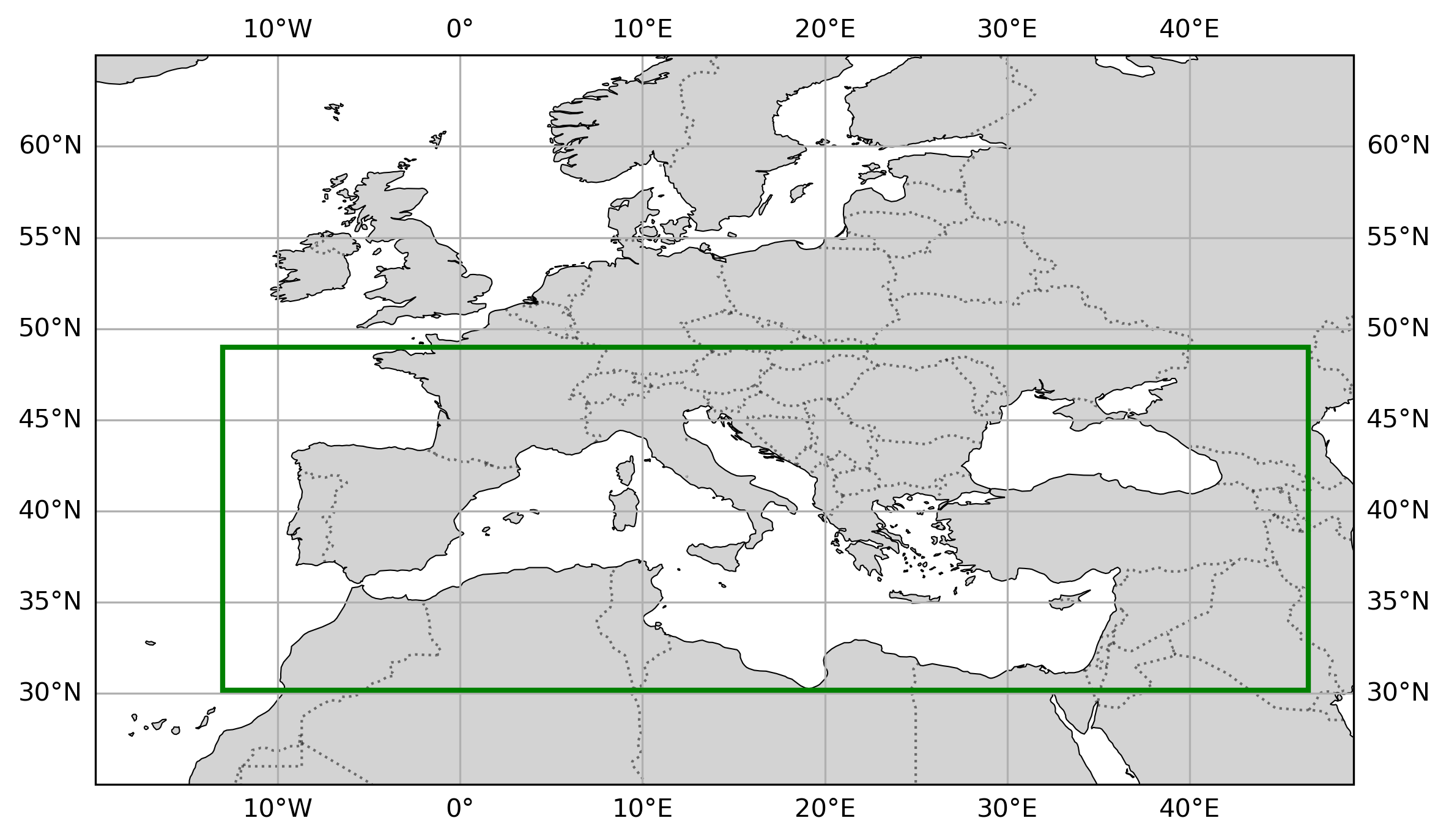}
    \caption{Area of Interest}
    \label{fig:fig2}
\end{figure}

\noindent \textbf{Fine-tuning datasets.} 
To evaluate our approach, we curated datasets for two downstream tasks: cloud segmentation and active fire detection. Both cover 2020–2024 over the Mediterranean and use 11-band SEVIRI radiances as input with pixel-level masks as targets. 

For cloud segmentation, SEVIRI scenes were paired with MODIS cloud masks (MOD35/MYD35) by matching acquisitions within 10 minutes, reprojecting MODIS labels onto the SEVIRI grid, and retaining only high-confidence cloud/clear-sky pixels based on MODIS quality flags. For fire detection, we generated fire masks from NASA FIRMS active fire detections (MODIS and VIIRS) collocated with SEVIRI acquisitions. To reduce false positives, detections were cross-validated with burned-area products from EFFIS\footnote{https://forest-fire.emergency.copernicus.eu/}  and FireHUB\footnote{http://ocean.space.noa.gr/diachronic\_bsm/}, and unsupported detections were discarded.

To prevent spatiotemporal leakage, we use a temporally disjoint split: 2020–2021 for training, 2022 for validation, and 2023–2024 for testing. For fire training, we keep only samples containing at least one fire pixel, while validation and test sets preserve the natural class distribution.



\subsection{Pretraining with Adapted SatMAE}

\textbf{The SatMAE architecture.} Masked Autoencoders (MAEs) are self-supervised models that learn data representations by reconstructing masked portions of the input. They consist of an encoder that processes visible input tokens to produce a latent representation, and a decoder that reconstructs the original input from this representation. During training, a large fraction of the input (e.g., image patches) is masked, and only the unmasked patches are passed through the encoder, typically, a Vision Transformer (ViT)~\cite{DosovitskiyB0WZ21}. The decoder—also composed of Transformer blocks—then operates on a combination of encoded visible tokens and placeholders for masked tokens. Positional embeddings are added to all tokens to preserve spatial structure and enable the model to infer the correct locations of the missing patches during reconstruction.
The SatMAE~\cite{cong2022satmae} implementation extends the MAE framework to the domain of satellite imagery in order to handle multi-spectral and multi-temporal data. SatMAE retains the core masked autoencoding structure but the ViT backbone supports spectral encoding and multi-temporal inputs to account for the specific characteristics of remote sensing inputs.


\noindent \textbf{The adapted SatMAE architecture.}
In this work, we build upon the original SatMAE implementation, incorporating architectural adaptations proposed by \cite{girtsou2024mlphys} to better align with the characteristics of our data and the requirements of our work. A key modification concerns the treatment of temporal information. While the original SatMAE design omits fine-grained timestamp components (e.g., minutes and seconds), assuming they are irrelevant for slowly evolving phenomena such as vegetation or land cover, the approach in \cite{girtsou2024mlphys} retains this information. We adopt the same strategy, as our focus on real-time monitoring demands higher temporal resolution. In scenarios involving rapidly evolving events, such as wildfires, minute-level temporal cues can carry critical predictive signals that enhance the model's responsiveness and precision.

To further investigate the role of short-term temporal context during pretraining, we trained SatMAE under two temporal settings: (i) a \emph{single-timestep} variant, where each sample contains one SEVIRI acquisition, and (ii) a \emph{multi-timestep} variant, where each sample contains three SEVIRI acquisitions of the same patch randomly selected within the same hour. This design enables the model to learn both instantaneous representations and representations informed by intra-hour variability, which is particularly relevant for fast-changing atmospheric and fire-related dynamics.

For the training of SatMAE we used a base ViT model (approximately 90 million parameters) with a hidden size of 768. The model input consists of 32×32 pixel patches extracted from SEVIRI scenes. Due to the small size of these inputs—driven by the coarse resolution of SEVIRI imagery—we used a 4×4 token embedding structure. This configuration was chosen to preserve as much spatial detail as possible. Larger token sizes (i.e., fewer, coarser patches) would risk oversmoothing critical spatial patterns, which are essential for accurate segmentation in low-resolution satellite data.



\subsection{Fine-tuning}

We assessed the expressiveness and generalization capability of the representations learned during pretraining on downstream tasks chosen for their rapid spatiotemporal evolvement; cloud presence and active fire detection. Both tasks are formulated as binary semantic segmentation and the objective is to identify the presence of cloud/fire at the pixel level using SEVIRI satellite imagery. Each input sample consists of a 32×32 image patch with 11 spectral bands, accompanied by a corresponding target binary mask labeling each pixel.

For this, we retained the encoders from our pretrained Vision Transformer models and extended them with custom segmentation decoders tailored for dense prediction. The decoders reconstruct spatially resolved output maps from the latent feature representations, leveraging a series of transposed convolutions and residual connections to facilitate effective upsampling and contextual refinement. This design allowed the models to integrate multiscale information and produced precise, high-resolution masks.

Through this fine-tuning setup, we aimed to determine how well the pretrained encoders adapt to the spatial and spectral characteristics relevant for the downstream tasks, and whether pretraining on SEVIRI imagery offers benefits over training from scratch, initializing from generic ImageNet checkpoints, or using existing EO FMs.

\section{Experimental Setup}



This section outlines the experimental setting and implementation details used to evaluate our approach, the training configurations and evaluation metrics used. 

Table \ref{table:dataset_stats} summarizes the datasets used for both the pretraining and fine-tuning stages. For the downstream tasks, we include the number of image samples as well as pixel-level annotations to highlight class imbalance in both datasets—most notably for active fire detection, where positive (fire) pixels account for less than 0.4\% of all pixels across splits.



\begin{table}[h]
\centering
\footnotesize
\setlength{\tabcolsep}{0.5pt}
\caption{Dataset Statistics for the HighFM model}
\begin{tabular}{l@{\hskip 8pt}c@{\hskip 8pt}c@{\hskip 8pt}c@{\hskip 8pt}c}
\hline
\textbf{Split} & \textbf{Images} & \textbf{Background Pixels} & \textbf{Target Pixels} & \textbf{Target Ratio} \\
\hline\hline
\multicolumn{5}{c}{\textbf{Pretraining dataset}} \\
\hline
Train & 13.6M & -- & -- & -- \\
Val   & 5.3M & - & -- & -- \\
Test  & 1.3M & -- & -- & -- \\
\hline
\multicolumn{5}{c}{\textbf{Active Fires dataset}} \\
\hline
Train & 2140 & 2.18M & 7.77k & 0.00355 \\
Val   & 1099 & 1.12M & 3.69k & 0.00327 \\
Test  & 2873 & 2.93M & 11.1k & 0.00378 \\
\hline
\multicolumn{5}{c}{\textbf{Clouds dataset}} \\
\hline
Train & 551 & 502K & 62K & 0.11 \\
Val   & 1614 & 214.9K & 1.6M & 0.13 \\
Test  & 671 & 584M & 103M & 0.16 \\
\hline
\end{tabular}
\label{table:dataset_stats}
\end{table}



To align with realistic operational needs, we design our fine-tuning experiments around two complementary use cases. Each setting emphasizes a different trade-off between coverage and precision, and is optimized and evaluated accordingly. These directions were derived by solar-energy stakeholders who use cloud masks for forecasting and prefer conservative “worst-case” outputs (tolerating false positives) and fire-response stakeholders who require low false-positive rates for trust and operational usability. 

\noindent\textbf{(a) High-Recall Detection.}
This setting prioritizes capturing as many positive pixels as possible, tolerating increased false positives when missing clouds or fires is more costly (e.g., early warning). All models are trained with weighted cross-entropy to mitigate class imbalance, and performance is monitored using \textit{balanced accuracy}.

\noindent\textbf{(b) Precision-Oriented Localization.}
This setting emphasizes spatially accurate, high-confidence masks, targeting fewer false positives and better interpretability for decision support. Models are optimized with Dice loss, and performance is evaluated using positive-class \textit{IoU}.

Across experiments, the validation set is used to tune hyperparameters (class weights from [$(1,1)$, $(1,500)$, $(1,1000)$, $(1,2000)$, $(1,5000)$, $(1,10000)$] and augmentation), and to select the best checkpoint according to the monitored metric. Final results are reported on the test set.

We benchmark our architecture against a set of baselines to evaluate the effectiveness of our domain-specific pretraining for the two downstream tasks.

\noindent\textbf{(a) UNet from Scratch.}
This baseline uses a standard UNet~\cite{RonnebergerFB15} architecture trained from scratch on the downstream tasks. As a widely adopted convolutional model for semantic segmentation, it serves as a reference point for performance on dense prediction tasks using multispectral satellite data.

\noindent\textbf{(b) Vision Transformer (ViT) from Scratch.}
We train a ViT-Base model without any pretraining to assess how well a transformer can learn task-relevant features directly from the segmentation tasks alone. This baseline isolates the impact of architectural inductive biases without the influence of pretrained model weights.

\noindent\textbf{(c) Vision Transformer (ViT) pretrained on ImageNet.}
This baseline leverages a ViT-Base model pretrained on the ImageNet-1k classification task \footnote{https://huggingface.co/google/vit-base-patch16-224}. We fine-tune it on both cloud segmentation and fire detection to evaluate how well generic visual representations transfer to multispectral segmentation tasks, providing a comparison point for the domain-specific SatMAE pretraining.

\noindent\textbf{(d) Copernicus-FM.}
This baseline uses the Copernicus-FM foundation model \cite{wang2025unifiedcopernicusfoundationmodel}, pretrained on a large-scale, multimodal corpus of Copernicus Sentinel data spanning multiple sensors and spectral configurations. Copernicus-FM produces flexible, sensor-aware representations via metadata-conditioned dynamic weights. Fine-tuning Copernicus-FM on the downstream tasks allows us to evaluate the benefits of large-scale, domain-general EO pretraining compared to our domain-specific strategy.

\noindent\textbf{(e) Panopticon.}
We include Panopticon~\cite{waldmann_shah_2025_panopticon}, an any-sensor Earth observation foundation model pretrained using self-supervised learning across co-registered multi-sensor satellite imagery. Panopticon employs a transformer backbone with spectral and sensor-aware channel embeddings to support robust generalization across heterogeneous inputs. Panopticon provides a strong baseline for assessing how well general multisensor EO representations transfer to specific segmentation tasks.

\begin{table*}[h]
    \centering
    \footnotesize
    \setlength{\tabcolsep}{4pt}
    \begin{tabular}{c c c c c c}
        \hline
        \textbf{Method} & \textbf{Balanced Accuracy} & \textbf{IoU\textsubscript{no-cloud}} &
        \textbf{IoU\textsubscript{cloud}} & \textbf{Recall\textsubscript{no-cloud}} &
        \textbf{Recall\textsubscript{cloud}} \\
        \hline\hline
        \multicolumn{6}{c}{\textbf{Models trained with cross-entropy loss}} \\
        \hline
        U-Net\textsubscript{scratch} & $0.792 \pm 0.002$ & $0.618 \pm 0.005$ & $0.701 \pm 0.003$ & $0.743 \pm 0.014$ & $\mathbf{0.842 \pm 0.010}$ \\
        ViT-B/4\textsubscript{scratch} & $0.826 \pm 0.001$ & $0.677 \pm 0.002$ & $0.727 \pm 0.008$ & $\underline{0.827 \pm 0.018}$ & $0.826 \pm 0.019$ \\
        \hline
        ViT-B/4\textsubscript{ImageNet} & $0.819 \pm 0.002$ & $0.667 \pm 0.002$ & $0.713 \pm 0.008$ & $\mathbf{0.829 \pm 0.013}$ & $0.808 \pm 0.016$ \\
        Copernicus-FM & $0.825 \pm 0.001$ & $0.672 \pm 0.002$ & $0.730 \pm 0.006$ & $0.809 \pm 0.014$ & $\underline{0.840 \pm 0.015}$ \\
        Panopticon & $0.826 \pm 0.002$ & $0.674 \pm 0.003$ & $0.730 \pm 0.004$ & $0.815 \pm 0.013$ & $0.837 \pm 0.012$ \\
        HighFM\textsubscript{ST} (Ours) & $\underline{0.828 \pm 0.003}$ & $\underline{0.678 \pm 0.004}$ & $\underline{0.731 \pm 0.010}$ & $0.823 \pm 0.018$ & $0.833 \pm 0.021$ \\
        HighFM\textsubscript{MT} (Ours) & $\mathbf{0.831 \pm 0.002}$ & $\mathbf{0.683 \pm 0.004}$ & $\mathbf{0.737 \pm 0.008}$ & $0.823 \pm 0.023$ & $\underline{0.840 \pm 0.023}$ \\
        \hline
        \multicolumn{6}{c}{\textbf{Models trained with Dice loss}} \\
        \hline
        U-Net\textsubscript{scratch} & $0.793 \pm 0.003$ & $0.615 \pm 0.005$ & $0.712 \pm 0.002$ & $0.716 \pm 0.011$ & $\mathbf{0.871 \pm 0.007}$ \\
        ViT-B/4\textsubscript{scratch} & $\underline{0.824 \pm 0.002}$ & $0.670 \pm 0.005$ & $0.734 \pm 0.004$ & $0.797 \pm 0.015$ & $0.851 \pm 0.013$ \\
        \hline
        ViT-B/4\textsubscript{ImageNet} & $0.819 \pm 0.003$ & $0.662 \pm 0.006$ & $0.728 \pm 0.001$ & $0.791 \pm 0.015$ & $0.847 \pm 0.009$ \\
        Copernicus-FM & $0.819 \pm 0.004$ & $0.662 \pm 0.010$ & $0.728 \pm 0.010$ & $0.791 \pm 0.040$ & $0.848 \pm 0.034$ \\
        Panopticon & $0.820 \pm 0.005$ & $0.661 \pm 0.011$ & $\underline{0.736 \pm 0.002}$ & $0.772 \pm 0.026$ & $\underline{0.868 \pm 0.016}$ \\
        HighFM\textsubscript{ST} (Ours) & $\mathbf{0.829 \pm 0.002}$ & $\mathbf{0.681 \pm 0.004}$ & $0.731 \pm 0.008$ & $\mathbf{0.828 \pm 0.024}$ & $0.830 \pm 0.023$ \\
        HighFM\textsubscript{MT} (Ours) & $\mathbf{0.829 \pm 0.002}$ & $\underline{0.678 \pm 0.006}$ & $\mathbf{0.740 \pm 0.007}$ & $\underline{0.804 \pm 0.030}$ & $0.854 \pm 0.026$ \\
        \hline
    \end{tabular}
    \caption{Test set performance of cloud segmentation models trained without data augmentation. Results are reported as mean $\pm$ std over 5 runs.}
    \label{table:clouds_results}
\end{table*}

\begin{table*}[h]
    \centering
    \footnotesize
    \setlength{\tabcolsep}{4pt}
    \begin{tabular}{c c c c c c} 
        \hline
        \textbf{Method} & \textbf{Balanced Accuracy} & \textbf{IoU\textsubscript{no-fire}} &
        \textbf{IoU\textsubscript{fire}} & \textbf{Recall\textsubscript{no-fire}} &
        \textbf{Recall\textsubscript{fire}} \\
        \hline\hline
        \multicolumn{6}{c}{\textbf{Models trained with cross-entropy loss}} \\
        \hline
        U-Net\textsubscript{scratch} & $0.834 \pm 0.004$ & $0.866 \pm 0.007$ & $0.022 \pm 0.001$ & $0.867 \pm 0.007$ & $0.802 \pm 0.007$ \\ 
        ViT-B/4\textsubscript{scratch} & $0.883 \pm 0.006$ & $0.937 \pm 0.010$ & $0.048 \pm 0.006$ & $0.937 \pm 0.010$ & $0.829 \pm 0.019$ \\ 
        \hline
        ViT-B/4\textsubscript{ImageNet} & $0.856 \pm 0.010$ & $0.942 \pm 0.010$ & $0.049 \pm 0.007$ & $0.943 \pm 0.010$ & $0.769 \pm 0.027$ \\ 
        Copernicus-FM & $0.894 \pm 0.003$ & $0.943 \pm 0.005$ & $0.053 \pm 0.004$ & $0.943 \pm 0.005$ & $0.845 \pm 0.007$ \\ 
        Panopticon & $0.888 \pm 0.019$ & $0.928 \pm 0.013$ & $0.044 \pm 0.008$ & $0.929 \pm 0.013$ & $0.848 \pm 0.028$ \\ 
        HighFM\textsubscript{ST} (Ours) & $\underline{0.917 \pm 0.003}$ & $\underline{0.953 \pm 0.005}$ & $\underline{0.066 \pm 0.005}$ & $\underline{0.954 \pm 0.004}$ & $\underline{0.881 \pm 0.006}$ \\ 
        HighFM\textsubscript{MT} (Ours) & $\mathbf{0.925 \pm 0.002}$ & $\mathbf{0.960 \pm 0.006}$ & $\mathbf{0.079 \pm 0.009}$ & $\mathbf{0.961 \pm 0.006}$ & $\mathbf{0.890 \pm 0.008}$ \\ 
        \hline
        \multicolumn{6}{c}{\textbf{Models trained with Dice loss}} \\
        \hline
        U-Net\textsubscript{scratch} & $0.684 \pm 0.004$ & $\underline{0.996 \pm 0.000}$ & $0.272 \pm 0.005$ & $\mathbf{0.999 \pm 0.000}$ & $0.369 \pm 0.008$ \\ 
        ViT-B/4\textsubscript{scratch} & $0.709 \pm 0.028$ & $\underline{0.996 \pm 0.000}$ & $0.286 \pm 0.043$ & $\underline{0.998 \pm 0.000}$ & $0.419 \pm 0.055$ \\ 
        \hline
        ViT-B/4\textsubscript{ImageNet} & $0.719 \pm 0.047$ & $0.994 \pm 0.004$ & $0.258 \pm 0.063$ & $0.997 \pm 0.004$ & $0.442 \pm 0.098$ \\ 
        Copernicus-FM & $0.717 \pm 0.011$ & $\underline{0.996 \pm 0.000}$ & $0.313 \pm 0.010$ & $\mathbf{0.999 \pm 0.000}$ & $0.436 \pm 0.021$ \\ 
        Panopticon & $0.709 \pm 0.038$ & $\underline{0.996 \pm 0.000}$ & $0.283 \pm 0.056$ & $\underline{0.998 \pm 0.000}$ & $0.420 \pm 0.075$ \\ 
        HighFM\textsubscript{ST} (Ours) & $\underline{0.740 \pm 0.008}$ & $\underline{0.996 \pm 0.000}$ & $\underline{0.340 \pm 0.004}$ & $\underline{0.998 \pm 0.000}$ & $\underline{0.481 \pm 0.016}$ \\ 
        HighFM\textsubscript{MT} (Ours) & $\mathbf{0.748 \pm 0.014}$ & $\mathbf{0.997 \pm 0.000}$ & $\mathbf{0.352 \pm 0.005}$ & $\underline{0.998 \pm 0.000}$ & $\mathbf{0.497 \pm 0.027}$ \\ 
        \hline
    \end{tabular}
    \caption{Test set performance of fire detection models trained with data augmentation. Results are reported as mean $\pm$ std over 5 runs.}
    \label{table:fires_results}
\end{table*}

All Vision Transformer (ViT)-based models, including ours and the baselines, employ the same ViT-Base architecture with matched parameter counts and identical segmentation heads to ensure a fair comparison. Fine-tuning of pretrained models updates the entire network, including both the encoder and the segmentation head. Models are trained using the Adam optimizer with a cosine annealing learning rate scheduler, a batch size of $64$, a learning rate of $1e-4$, and for up to $150$ epochs.

\section{Results}


We evaluate the detection tasks under the two distinct training objectives: (i) maximizing positive class recall and (ii) producing spatially precise segmentation maps. Tables \ref{table:clouds_results} and  \ref{table:fires_results} report results for cloud segmentation and active fire detection, respectively. For each task, we separate experiments by loss function and corresponding model-selection criterion: cross-entropy, with selection based on validation balanced accuracy to prioritize recall, and Dice loss, with selection based on validation IoU to emphasize spatial precision. In both settings, we train models with and without data augmentation and benchmark our HighFM variants—single-timestep (HighFM\textsubscript{ST}) and three-timestep multi-temporal (HighFM\textsubscript{MT})—against the selected baselines. In the main tables, we report only the best-performing configuration for each of our models; the full ablation results are provided in the Appendix. We observe that augmentation benefits active fire detection—consistent with fires occupying very small, sparse regions—whereas clouds typically cover larger portions of each patch and do not consistently gain from augmentation.

Table \ref{table:clouds_results} compares cloud segmentation performance on the test set, with no data augmentation applied during training. Among models trained with cross-entropy loss, the best overall performance is achieved by HighFM\textsubscript{MT}, which attains the highest balanced accuracy (0.831) and the best IoU for both no-cloud (0.683) and cloud (0.737), indicating improved separation of clear-sky and cloudy pixels. Under Dice loss, our models are also among the strongest performers: both achieve the highest balanced accuracy (0.829), while HighFM\textsubscript{MT} yields the highest cloud IoU (0.740) and HighFM\textsubscript{ST} provides the strongest no-cloud IoU (0.681) and no-cloud recall (0.828). Although U-Net and Copernicus-FM achieve slightly higher cloud recall in some settings, this comes at the expense of lower no-cloud recall, leading to reduced balanced accuracy and lower IoU overall. Finally, we do not observe a consistent advantage of one loss function over the other: performance is broadly comparable across cross-entropy and Dice, suggesting that despite class imbalance the models learn cloud patterns reliably.

The results in  Table \ref{table:fires_results} demonstrate that all models trained with cross-entropy loss are capable of detecting fire regions to varying extents. In contrast to cloud segmentation, prioritizing fire recall in this setting can sometimes result in over-segmentation, as reflected in the relatively low fire IoU scores across all models trained with cross-entropy. This indicates that while most fire occurrences are detected, the predicted regions may overestimate the actual fire extent. Within this context, our HighFM\textsubscript{MT} model consistently outperforms all the baselines. It achieves the highest balanced accuracy (0.925) and fire recall (0.890), surpassing slightly our HighFM\textsubscript{ST} and the next best performing baseline model (Copernicus-FM) by +0.031 and +0.045 respectively. 
These improvements are significant in the context of critical applications such as early wildfire detection. 
In the case of Dice loss training, which emphasizes spatial precision, all models produce more accurate and concentrated fire segmentations, as indicated by higher fire-class IoU values compared to the cross-entropy setting. We observe that this comes at the cost of reduced fire-class recall, with models becoming more conservative and occasionally failing to detect some fire incidents with limited spatial patterns. As with the previous objective, our proposed HighFM\textsubscript{MT} model outperforms all baselines, achieving the highest fire IoU of 0.352
, exceeding the best baseline (ViT-B/4\textsubscript{ImageNet}) by +0.039, while also attaining the highest fire recall of 0.497. These results confirm our model’s ability to deliver spatially precise and complete active fire delineations without compromising detection performance.


\begin{figure*}[h!]
    \centering
    \includegraphics[width=\textwidth]{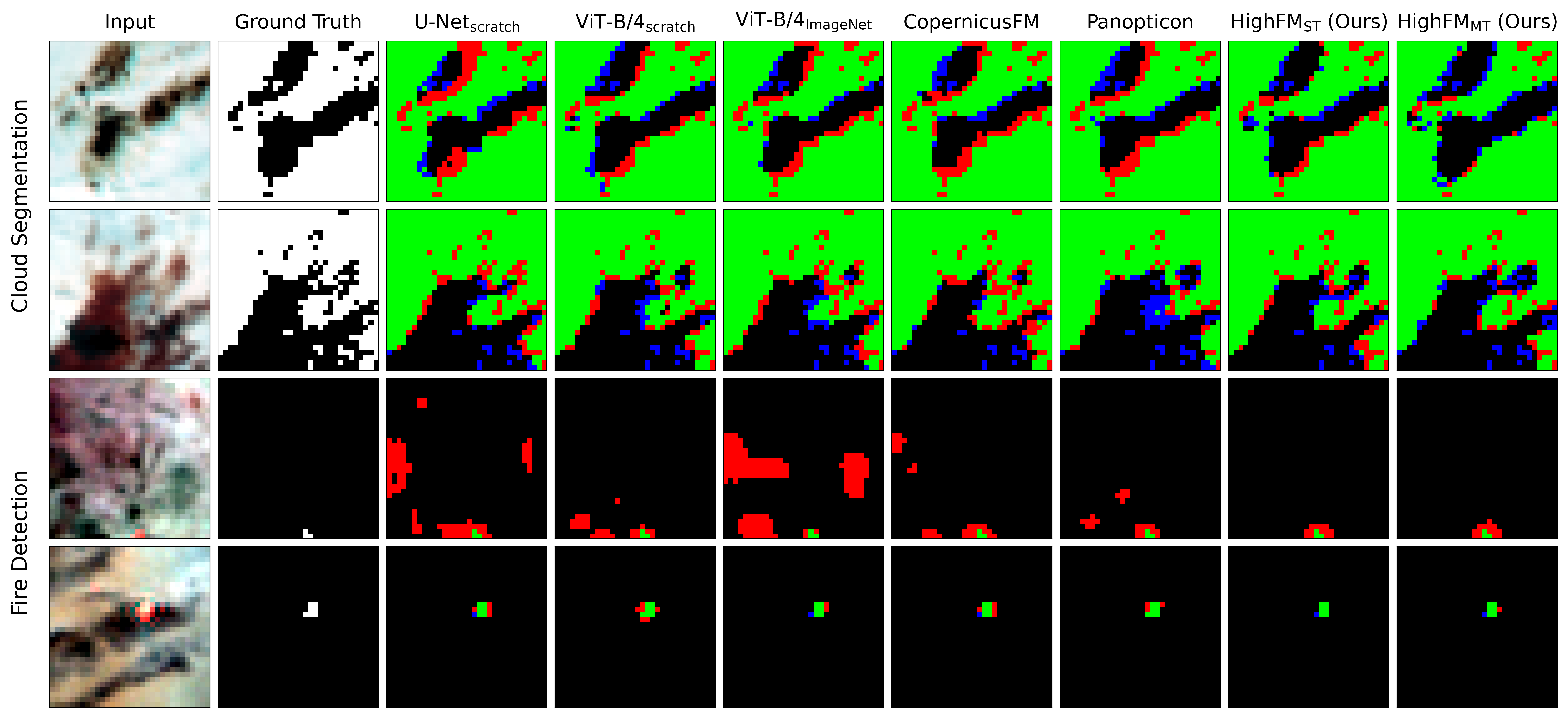}
    \caption{Qualitative results on test samples. Rows 1 and 2 show models fine-tuned on the cloud segmentation task, while rows 3 and 4 correspond to the fire detection task. Rows 1 and 3 present outputs from models trained with cross-entropy loss, whereas rows 2 and 4 show results from models trained with Dice loss. SEVIRI input images are visualized using a false-color fire composite, consistent with EUMETSAT’s operational fire RGB products. In the prediction columns, true positives are shown in green, false positives in red, and false negatives in blue.}
    \label{fig:qualitative_results}
\end{figure*}

In both tasks, our SEVIRI-pretrained model consistently outperforms baselines, demonstrating the benefits of leveraging domain-specific pretraining. In the fire segmentation task, across both training objectives, the trade-off reflects a shift from broad detection coverage to fine-grained localization, aligning with different operational priorities.
These results highlight the adaptability of the proposed approach in supporting different application needs, whether prioritizing early detection through high fire recall or enabling precise intervention via accurate segmentation.



Figure \ref{fig:qualitative_results} presents sample qualitative results for the downstream tasks under the two distinct use cases: models trained with cross-entropy loss, optimized for higher recall, and models trained with Dice loss, focused on producing more precise predictions. The main differences are observed in fire detection task; in the cross-entropy setting, the proposed model consistently detects fire incidents, including challenging cases with minimal signal, while introducing significantly fewer false positives. Its outputs are compact and well-localized, in contrast to baseline models, which tend to oversegment and activate large, irrelevant regions, indicating higher sensitivity to noise and limited spatial discrimination. In the Dice loss setting, the focus shifts toward achieving higher precision. In this context, the proposed model again demonstrates superior performance, producing accurate fire segmentations with minimal over- or under-segmentation. Compared to the baselines, it results in fewer false positives and improved boundary alignment with the ground truth fire masks. While all models display more conservative behavior under Dice training, the proposed model stands out in its ability to retain true positives without introducing false detections. These results highlight the value of combining our domain-specific ViT backbone with task-appropriate training objectives: enabling the model to either robustly detect subtle fire patterns under recall-oriented settings (cross-entropy) or provide fine-grained segmentations when precision is prioritized (Dice).


\section{Discussion}


In this work, we lay the groundwork for developing a foundation model (FM) tailored to real-time monitoring of fast-evolving physical phenomena from satellites.  Leveraging data from the geostationary MSG/SEVIRI sensor, we demonstrate that domain-specific pretraining significantly enhances performance, despite the relatively coarse spatial resolution of the imagery. The high temporal resolution, combined with extensive pretraining, enables our model to perform robustly in real-time settings. Our approach consistently outperforms all baseline models, highlighting the importance of temporal density and domain adaptation in EO FMs.
Furthermore our model is scalable and reusable for real-time EO tasks beyond cloud and fire detection. The same model can be used for a variety of downstream tasks like nowcasting of severe weather phenomena or solar energy forecasting — all of which demand rapid, continuous observation at large scales.
Real-time EO models, such as the one proposed in this work, have growing importance in societal and civil protection contexts. Timely alerts based on trustworthy systems that have been trained with an abundance of EO data are critical for informed decision-making and operational response. Our model contributes to the development of cutting-edge EO systems capable of supporting civil protection agencies, environmental monitoring services, and climate resilience initiatives.
This initial implementation focuses on a single sensor to isolate and validate the benefits of this approach. However, real-time EO landscape is increasingly multimodal, with sensors such as MODIS and VIIRS playing critical roles in EO tasks. Current single-modality pretraining restricts the model's ability to generalize across sensors and resolutions. Future work will address this limitation by exploring multimodal pretraining approaches with resolution-agnostic pretraining strategies, aiming to integrate heterogeneous satellite data sources into a unified, robust real-time monitoring framework.

\bibliographystyle{named}
\bibliography{ijcai26}

\end{document}